\documentclass[10pt,journal]{IEEEtranTCOM}
\usepackage{times,amsmath,amssymb,graphicx,epsfig,color,colortbl}
\usepackage{tikz}
\usepackage{caption}
\usepackage{soul}
\usepackage[normalem]{ulem}
\usepackage{xcolor}
\usepackage{algorithm}
\usepackage[noend]{algpseudocode}

\makeatletter
\def\BState{\State\hskip-\ALG@thistlm}
\makeatother

\def\conf{\text{conf}}

\def\Prob{\text{Prob}}
\def\Pred{\text{Pred}}
\def\and{\text{and}}

\def\mb#1{\mathbf{#1}}

\def\nn{\nonumber}
\def\beq{\begin{equation}}
\def\eeq{\end{equation}}
\def\beqa{\begin{eqnarray}}
\def\eeqa{\end{eqnarray}}

\def\nn{\nonumber}

\def\beq{\begin{equation}}
\def\eeq{\end{equation}}
\def\beqa{\begin{eqnarray}}
\def\eeqa{\end{eqnarray}}

\def\bmtx{\begin{bmatrix}}
\def\emtx{\end{bmatrix}}

\begin{document}

\title{\bf Code Failure Prediction and Pattern Extraction \\using LSTM Networks}
\author{Mahdi Hajiaghayi and Ehsan Vahedi$^{*}$\thanks{$^{*}$Mahdi Hajiaghayi is a Senior Machine Learning Scientist with the AI+Research group at Microsoft, Ehsan Vahedi is a Senior Data Scientist with Microsoft Office and an IEEE Senior Member.
(c)2018 Microsoft Corporation. All rights reserved.  This document is provided "as-is." Information and views expressed in this document, including URL and other Internet Web site references (if any), may change without notice. You bear the risk of using it. This document does not provide you with any legal rights to any intellectual property in any Microsoft product.}\\
\IEEEauthorblockA{(c)Microsoft Corporation, Redmond, WA 
\\E-mail: \{mahajiag, ehvahedi\}@microsoft.com}}
\maketitle

\thispagestyle{empty}
\pagestyle{empty}

\begin{abstract}
In this paper, we use a well-known Deep Learning technique called Long Short Term Memory (LSTM) recurrent neural networks to find sessions that are prone to code failure in applications that rely on telemetry data for system health monitoring. We also use LSTM networks to extract telemetry patterns that lead to a specific code failure. For code failure prediction, we treat the telemetry events, sequence of telemetry events and the outcome of each sequence as words, sentence and sentiment in the context of sentiment analysis, respectively. Our proposed method is able to process a large set of data and can automatically handle edge cases in code failure prediction. We take advantage of Bayesian optimization technique to find the optimal hyper parameters as well as the type of LSTM cells that leads to the best prediction performance. We then introduce the \textit{Contributors} and \textit{Blockers} concepts. In this paper, contributors are the set of events that casue a code failure, while blockers are the set of events that each of them individually prevents a code failure from happening, even in presence of one or multiple contributor(s). Once the proposed LSTM model is trained, we use a greedy approach to find the contributors and blockers. 
To develop and test our proposed method, we use synthetic (simulated) data in the first step. The synthetic data is generated using a number of rules for code failures, as well as a number of rules for preventing a code failure from happening. The trained LSTM model shows 
over $99\%$ accuracy for detecting code failures in the synthetic data. The results from the proposed method outperform the classical learning models such as Decision Tree and Random Forest. Using the proposed greedy method, we are able to find the contributors and blockers in the synthetic data in more than $90\%$ of the cases, with a performance better than sequential rule and pattern mining algorithms. In the next step, we train and test our proposed LSTM method on real data that we collected from sequences of activities performed by millions of Microsoft Office customers.

\end{abstract}

\noindent\textit{\textbf{Keywords \textemdash}} Deep learning, LSTM, Bi-LSTM, bayesian optimization, sequential rule mining, random forest, telemetry data, crash and code failure predication.

\section{Introduction}

Identifying the set of components and attributes that result in (or contribute to) code failure is an important topic in any application that relies on telemetry data for system health monitoring. For example, in Microsoft Office products, test engineers are interested in finding a generic pattern in the data 
that causes a code failure. Usually this is not an easy problem to solve since a combination of many factors (such as user's activities, hardware architecture, operating system, other programs running in the background, add-ins, etc) can potentially contribute in a code failure. Also part of this information may not be fully captured by telemetry signal. Moreover, a code failure may not be necessarily tied to the very last activity of the user, but triggered by a sequence of activities with specific order. Also depending on the architecture design, we might capture or lose the very last batch of telemetry data if a major failure such as crash happens. 

In this paper, we are interested in finding the root-cause of code failures as well as building a model to predict the code failure.
We use Long Short Term Memory (LSTM) which is a type of recurrent neural networks for code failure prediction and pattern extraction.

The novelty of this work can be summarized as follows: 
\begin{itemize}
\item For code failure prediction, we propose an LSTM network whose hyper-parameters and type of LSTM cells are determined by Baysian optimization technique \cite{snoek2012practical}. It is usually not a trivial task to pick the LSTM architecture that achieves the best performance \cite{greff2017lstm}. Bayesian optimization technique enables us to systematically find the best LSTM architecture for our application. In this work, we consider two types of LSTM networks: (a) standard, and (b) Bidirectional network. These topics are covered in Section~\ref{Prediction_model}

\item For code failure pattern extraction, we first introduce the \textit{Contributors} and \textit{Blockers} concepts. In this paper, contributors are the set of actions or events that individually or together result(s) in a code failure, while blockers are the set of actions or events that individually or together prevent(s) a code failure from happening. We then formulate the problem of finding contributors and blockers as two optimization problems and propose an algorithm that utilizes a trained LSTM-based prediction model to extract contributors and blockers in sequential data. Details of the proposed method are discussed in Section~\ref{pattern_extraction}.

\item In Section~\ref{experimental_work}, we provide experimental results that show the proposed method outperforms the existing algorithms in achieving better pattern extraction and prediction performance.  
\end{itemize}

\section{Related Works}

For sequential data analysis and prediction, conventional machine learning algorithms such as SVM, Logistic Regression and Feed-Forward Neural Networks assume independence between features and exhibit poor performance when the order of components in a sequence is important. In the Bag of Words, we lose the order of components in the sequences and although we can capture the order indirectly by creating $n$-grams and considering a sliding window, this type of solutions are not quite useful in time series analysis.   
Hidden Markov Models (HMM) and high order Markov chains can be another option for representing time series and an alternative for sequential modeling \cite{KINNEBREW2016Markov,scholz2016Markov,Melnykov2016Markov,Jurafsky2017Markov}. However, the state space grows exponentially with the size of the sequence, window size and number of states, rendering markov models computationally impractical for modeling long-term dependencies \cite{lipton2015critical}. 
Classic sequential pattern \cite{zaki2001spade} \cite{fournier2014fast} and rule mining methods \cite{lo2009} \cite{fournier2015mining} can help us gain insight about events that contribute to code failures when the size of data is not very large and the overall length of sequences or the window size is relatively small. However, these methods fail to handle long sequences and large datasets. The main drawback of these methods is the number of patterns and rules they need to keep track of, similar to Markov models growing exponentially with the length of sequences. Also sequential rule mining solutions perform well on sequences with no duplicate events, but this assumption (having no duplicate) is not necessarily valid for code failures we see during sessions conducted by Microsoft Office users and we cannot eliminate the possibility of having duplicates in the sequences of telemetry events.

The main advantage of Recurrent Neural Networks (RNN) and in particular LSTM networks is that they are end-to-end differentiable with respect to each of the parameters in the model. Therefore, unlike the other models with combinatorial nature, RNNs and LSTM networks can be trained using gradient-based algorithms. In fact, the training complexity of LSTM grows linearly with the number of weights in the network \cite{hochreiter1997long}, which makes LSTM networks very efficient for this type of problems. Moreover, an RNN model can avoid over-fitting using standard techniques such as drop-out and weight decay.

In addition to Natural Language Processing (NLP), LSTM networks have been used for many other applications including but not limited to time-series anomaly detection \cite{malhotra2015long}, speech recognition \cite{graves2013speech} and human action recognition \cite{baccouche2011sequential}. Broader architectural revisions of LSTM networks have also been proposed in recent years. Bidirectional \cite{graves2013hybrid}, multi-layer and recursive tree-structures \cite{socher2014recursive} are some examples of LSTM revisions.

Code failure prediction in sequential telemetry data is one of the areas that LSTM networks can offer an edge over classical machine learning and data mining methods. Zhang et al. recently used standard LSTM networks for system failure prediction \cite{ZhangFailurePrediction2016}, however, the focus of their work is on failure prediction and not identifying the root-cause of failures. We are, on the other hand, more interested in identifying the root-cause of code failures and extracting patterns that lead to code failures, in addition to code failure prediction.

Interpretability is critical for some applications such as medical diagnostic tools and self driving cars, where the reliance of the model on the correct features needs to be guaranteed \cite{Montavo2018,Caruana2015}. 
Recently there have been some efforts by researchers to better understand the decision making process in recurrent neural networks with applications in NLP \cite{li2016understanding} \cite{li2015visualizing} \cite{karpathy2015visualizing} \cite{lei2016rationalizing} and Genomic Sequencing \cite{lanchantin2017deep}. 
Karpathy et al. visualized the neural generation models from an error-analysis point of view, by analyzing predictions and errors of recurrent neural networks \cite{karpathy2015visualizing}. The approach shows the intriguing dynamics of hidden cells in LSTM networks, but is limited to a few manually-inspected cases such as brace opening and closing. Li et al. used the first-order derivative to examine the saliency of input features \cite{li2015visualizing}, but they relied on the overly strong assumption that the decision score is a linear combination of input features. 
In another effort, Li et al. trained a separate generator that extracts a subset of text which leads to a similar decision as the one with the original input, and used it to form an interpretable summary \cite{li2016understanding}. 
Lei et al. proposed a learning process that generates rationales for a given text \cite{lei2016rationalizing}. Rationales are subsets of words from the input text that are sufficient for prediction, and can be used as a substitute of the original text. 

Our work is different from \cite{lei2016rationalizing} in the sense that it extracts a set of words that may not be coherent, and can consist of words that are not necessarily in immediate sequences. The emphasis of our work is on the importance of each individual word in the sequence. Here the idea is to calculate the relative change of score for a text when a word is removed from the sequence. We utilize this approach to find the sequences of actions that lead to a specific code failure.

\section{Prediction Method} \label{Prediction_model}

\subsection{Network Architecture}

\begin{figure}[h]
 \centering
 \includegraphics[scale=1]{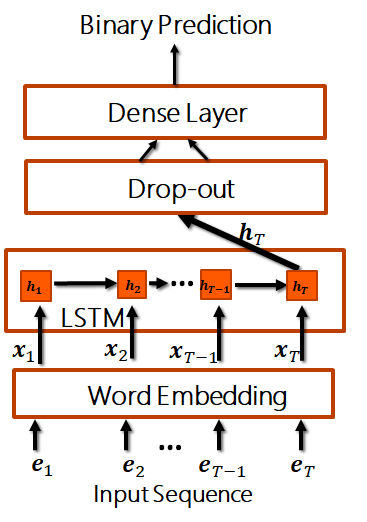}
 \caption{Proposed network architecture for code failure prediction}
 \label{network}
 \end{figure}

Suppose a session consists of a sequence of events that may or may not result in a code failure. Using a dataset of such sequence data, we train an LSTM-based model to predict the outcome of sequences (code failure vs no code failure). Fig. \ref{network} shows the high level architecture of our proposed LSTM model. First, the sequence of events $e_1, \cdots e_T$ is fed to the embedding layer where each event is encoded to an $n$-dimensional real-valued vector $\mb{x}_t$. Here $n$ is a parameter which is determined based on the number of events and the size of the input sequence. The output of the embedding layer is passed through the LSTM layer followed by a dropout layer to avoid over-fitting. In the end, we have a fully connected (Dense) layer that generates a single output value. In this setting, output $0$ indicates ``no code failure" while $1$ indicates ``code failure". 

For training, we use the Cross-Entropy loss function \cite{kevin2012machine}. Given the ground-truth vector of outcomes $\mb{y} = [y_1, \cdots, y_N]$ and the vector of outcomes predicted by our model $\mb{\hat{y}} = [\hat{y}_1, \cdots, \hat{y}_N]$ for $N$ sequences of data, the cross-entropy loss function is calculated as 

\beq
loss(\mb{y},\mb{\hat{y}}) = \frac{1}{N} \sum_{i=1}^{N} y_i \log(\frac{1}{\hat{y}_i}) + (1-y_i) \log(\frac{1}{1-\hat{y}_i}).
\eeq

\subsection{Long Short Term Memory (LSTM)}

 \begin{figure}[h]
 \centering
 \includegraphics[scale=.7]{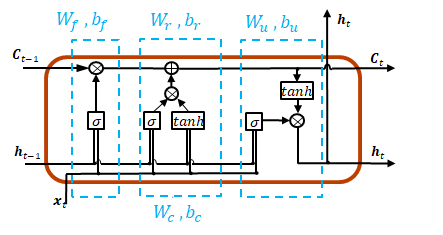}
 \caption{Components of a standard LSTM cell with corresponding weights and biases. The picture is partially borrowed from \cite{Olah2015lstm}.}
 \label{LSTMlayer}
 \end{figure}

LSTM is a special type of RNNs, capable of learning both \textit{long-term} and \textit{short-term} dependencies in data. An LSTM network consists of multiple LSTM cells. Each LSTM cell has three main components responsible for \textit{forgetting}, \textit{remembering} and \textit{updating} data \cite{hochreiter1997long}. These components are depicted in Figure~\ref{LSTMlayer}. For a time step $t$ and at cell $t$, we have the following input-output relationships: 

\begin{align}
\mb{f}_t & = \sigma(\mb{W}_f[\mb{h}_{t-1}, \mb{x}_t] + \mb{b}_f)  \nn \\
\mb{r}_t &= \sigma(\mb{W}_r[\mb{h}_{t-1},\mb{x}_t] + \mb{b}_r) \nn \\
\tilde{\mb{C}_t} &= \tanh(\mb{W}_u[\mb{h}_{t-1},\mb{x}_t] + \mb{b}_u) \nn \\
\mb{C}_t &= \mb{f}_t \odot \mb{C}_{t-1} + \mb{r}_t \odot \tilde{\mb{C}_t} \nn \\
\mb{o}_t &= \sigma(\mb{W}_o[\mb{h}_{t-1},\mb{x}_t] + \mb{b}_o) \nn \\
\mb{h}_t &= \mb{o}_t \odot \tanh(\mb{C}_t), \label{lstm}
\end{align}
where $\mb{x}_t \in \mathbf{R}^n$ is the vector representing input event at time $t$ and $\sigma(.)$ stands for the sigmoid function. In this notation, $\odot$ denotes the element-wise product. $\mb{h}_t$ and $\mb{h}_{t-1} \in \mathbf{R}^l$ are the output values of the hidden layers at time $t$ and $t-1$, respectively. $\mb{h}_t$ can also be viewed as the filtered version of cell state $\mb{C}_t$. While $\mb{f}_t$ adjusts how much activation is added to the internal state (forget gate), $\mb{o}_t$ controls the effect of the internal state on the next cell (output gate). $\mb{r}_t \odot \tilde{\mb{C}_t}$ is responsible for remembering and updating the input value $\mb{x}_t$.
The coefficients  $\mb{W}_i \in \mathbf{R}^{l+n};  \quad  i\in \{f,r,o,u\}$ and bias factors $\mb{b}_i \in \mathbf{R}^{l}; \quad  i\in \{f,r,o,u\}$ are optimization parameters shared among all cells. In total, there are $4(l+n) + 4l$ parameters to optimize for a standard LSTM network. In this paper, we also consider Bi-directional LSTM networks. Bi-directional LSTM networks can be thought of as two attached standard LSTMs with forward and backward directions. While the forward direction effectively makes use of the past features, the backward direction utilizes the forward features. For such networks, the hidden layer at time $t=1$ is given by $\mb{h}_T = [\overrightarrow{\mb{h}_T},\overleftarrow{\mb{h}_T}]$. Subsequently, a bi-directional LSTM network has twice as many variables as a standard LSTM. These parameters are determined during the training process. In addition to LSTM, we also tried other types of networks such as GRU and simple RNN for this application, but LSTM had better performance than GRU and RNN.

Feature engineering can be very challenging in some applications. An important advantage of using LSTM networks is that they do not require any feature engineering. Moreover, it needs no prior knowledge of the events that form a sequence or session, and all important features are identified by the algorithm itself.

\subsection{Bayesian Optimization For Hyper-parameter Tunning}
Any LSTM model has some hyper-parameters that need to be set before the training process starts. These hyper-parameters include the size of embedding layer ($n$), size of hidden layer ($|\mb{h}_T|$), learning rate and the type of LSTM network (standard vs. bi-directional).   
We use Bayesian optimization technique \cite{snoek2012practical} to find the optimal values of hyper-parameters for our model. It uses a probabilistic model for the objective function, which is the performance of the learning algorithm in this application, and based on the probabilistic model determines the next point to evaluate the function. The idea is to use all the available information to decide the next point, as opposed to only relying on the last point in conventional gradient methods. It is particularly appealing when the objective function is hard or expensive to calculate, with a good example being a deep neural network. This technique finds a nice trade-off between exploration and exploitation, and picks the hyper-parameters of the next iteration based on minimizing an accusation function. In our application, we used expected improvement (EI) \cite{Mockus2014} as the accusation function. The F1-score in $5$-fold cross-validation was set as the objective function of the optimization algorithm.  
Our model's hyper-parameters include learning rate, embedding size, and LSTM type (standard vs bi-directional cells). Using Bayesian optimization technique, we found the following optimal values of parameters for our synthetic dataset: 
embedding size ($N) = 3$, learning rate $= 0.02$, LSTM size$= 6$ and LSTM type = bi-directional.  We observed that most of the off-the-shelf embedding layers with embedding size of $20$ and more would lead to over-fitting problem. 

\section{Pattern Extraction} \label{pattern_extraction}
LSTM-based models are very good in predicting the outcome of a given sequence, however, they are often hard to interpret. 
In code failure prediction, it is important for the test engineers to find the root-cause of failures. More specifically, it is important to extract and identify patterns and combination of events that either result in code failure (contributors) or prevent a code failure from happening (blockers) during a session. 
For a given sequence of events $S= [e_1,\cdots, e_T]$ that leads to a code failure, the contributors $C$ are formally derived from the following optimization problem  

\begin{align}
\min_{C \subset S}& |C| \nn \\
s.t  \quad &\Pred(S) = 1  \nn \\ 
&\Pred(S\setminus c_i) = 0, \quad \forall c_i \in C 
\end{align} 

\noindent where $S\setminus c_i$ refers to removing event $c_i$ from $S$. Here, we assume that a set of events collectively contribute to a code failure and if one of these events is removed we will not see the code failure. 

Similarly, blockers $B$ are identified by solving the optimization problem below 

\begin{align}
\min_{B \subset S}& |B| \nn \\
s.t  \quad &\Pred(S) = 0  \nn \\ 
&\Pred(S\setminus B) = 1, 
\end{align} 

\noindent where $S\setminus B$ refers to removing blocker $B$ from $S$. This notation is based on the assumption that each blocker by itself can prevent the code failure from happening. Here, it is assumed that sequence $S$ ends with no code failure (executed normally).

A naive approach to solve these two optimization problems and find blockers and contributors is to use an exhaustive search where all combinations of actions are examined to find the one with the minimum length that satisfies the constraints. However, the search space exponentially grows in this case, rendering the exhaustive search infeasible. 
In what follows, we propose a greedy approach that works well in identifying the contributors and blockers. 
For a given sequence $S$, we start from left to right and remove each event from the sequence. After each removal, we evaluate the output of the prediction model. If the prediction changes from $1$ to $0$, we keep the event in the contributors list, and if the prediction changes from $0$ to $1$ we add it to the blockers list. Removing events can be done in two ways. In the first approach which is called \textit{zero-inserting}, we replace the event under inspection with a default ``don't care" event, for example $0$. The second approach is called \textit{void-inserting} where we completely remove the event under inspection and make the sequence shorter. As it was not obvious which approach is better, we empirically examined both approaches. Based on the results which will be discussed in detail in the next section, we observed that void-inserting approach outperforms zero-inserting method.  
 

The approaches described above fail to detect the correct contributors and blockers in a sequence with duplicates. For example, consider $S = [a,f,b,h,b,c,j,f]$ and assume that we know actions $[f,b,c]$ are the contributors. If we remove any of the two $b$ events, the predicted label still remains the same due to the other event $b$ in the sequence. To address this problem, we need to remove the first and second $b$ events together, unless the first one has already been detected as a contributor or blocker. With this modification, the whole algorithm is described in \ref{algo1}.

\begin{algorithm} 
\caption{Contributor and blocker extraction algorithm for sequence $S$}\label{algo1}
\begin{algorithmic}[1]
\State Let $\Prob(S)$ = Confidence score for code failure likelihood of sequence $S$
\State $|S| = L$
\State Contributors: $C = \{\}$, Blockers : $B=\{\}$

\If {$Prob(S) > \conf_{th}$}
	\For {$k \gets 1$ to $L$} 
		\State $M \gets S$
		\For {$j \gets 1$ to $k$}
			\If {$S[j]=S[k] \quad \and \quad S[j]\notin B \cup C$} 
				\State $ M  \gets {M} \setminus {S}[j]$
			\EndIf
		\EndFor
		
		\State $diff = \Prob({M}) - \Prob({S})$
		\If {$|diff|$ $>$ $diff_{th}$}
			\If {$diff < 0$}
				\State  $ C.insert(S[k]) $
			\Else
				\State $ B.insert( S[k]) $
			\EndIf
		\EndIf

	\EndFor
\EndIf
\Return $B$ and $C$

\end{algorithmic}
\end{algorithm}

Another approach to solve the code failure prediction problem is to use sequential rule mining and sequential pattern mining techniques. Sequential rule mining algorithms \cite{zaki2001spade} \cite{fournier2014fast} \cite{fournier2011rulegrowth} discover rules in the form of $X \Rightarrow Y$ in a sequence of database such that $X$ and $Y$ are sequential patterns. Each rule is given by its support, which is the frequency of sequences that contain the rule, and confidence, which is the likelihood of sequence $Y$ appearing after $X$. The input of these algorithms are often \textit{minsup} (minimum support) or \textit{minconf} (minimum confidence) such that only rules whose support or confidence are higher than these threshold values are returned. This dependence on minsup and minconf thresholds could be a drawback of sequential rule mining algorithms for applications such as code failure prediction, since the threshold values are not known a priori. Moreover, as we see in the experimental section, while sequential rule mining methods can extract the contributors in our application, they are not able to detect blockers. Finally, 
the search space grows exponentially with the size of the database and the number of distinct actions in sequential rule mining algorithms. Based on the above, sequential rule mining algorithms are not capable of solving the code failure precition problem efficiently.

  
 

\section{Experimental Work} \label{experimental_work}

\subsection{Code failure Prediction using Synthetic Data}
In order to better understand how accurately and efficeintly our proposed LSTM model can solve the code failure prediction problem, we first applied our LSTM model on synthetic (simulated) data. The synthetic data was generated by assuming that we have $20$ distinct events $\{a, b, c, d, e, f, g, h, i, j, k, l, m, n, o, p, q, r, s, t\}$ and each session can have any combination of these $20$ events with a fixed length of $15$ events per session. The order of events are important (for example, the sequence $\{a, b\}$ is different from the sequence $\{b, a\}$). We assumed that any session containing the $\{f, b, c\}$ sequence will lead to code failure unless event $\{e\}$ is seen in the sequence. For example, a session with $\{a, b, b, f, b, c, j, d\}$ sequence of events will result in code failure but another session with $\{a, b, b, f, e, b, c, j, d\}$ will not result in any code failure. We randomly generated $30,000$ sequences using the above rule and split them $50/50$ for training and test sets. 
In addition to the proposed LSTM model, we used Decision Tree and Random Forest algorithms and compared the results from these three methods to see which one can better predict a code failure based on the simple rules we defined above.

\subsection{Code Failure Detection using Real Data}
In the next step, we trained our proposed LSTM model using real data from Excel users, which consists of Microsoft Office customers who paid to use our services. We removed all the identifiers from data to proetct users privacy. To make this dataset, we extracted all telemetry events recorded from our Excel users during the sessions they used Excel. We balanced the dataset to have one quarter with label 0 (code failure) and the rest with lable 1 (no code failure) data. In the next step, we trained an LSTM model that can detect and predict a specific type of code failure using the following parameters:
\noindent epochs: 1000, drop-out: 0.4, maximum sequence length: 45, embedding size: 2, training-test ratio: 50/50, number of sessions: 90,000
number of distinct telemetry events: 1067.

\subsection{Results and Performance Comparison}
\begin{figure}[h]
 \centering
 \includegraphics[scale=.6]{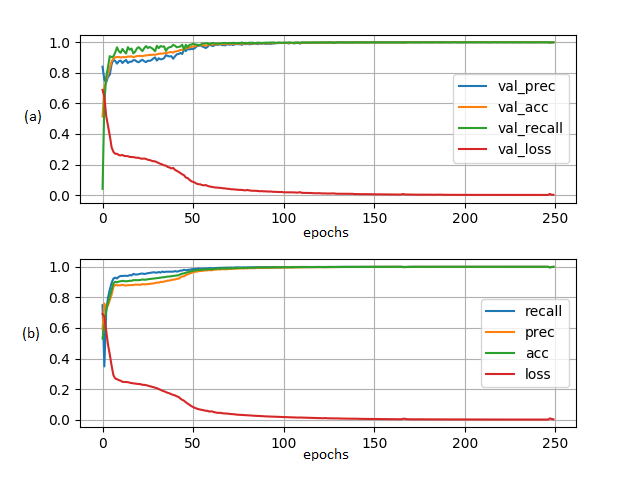}
 \caption{Performance of our proposed bi-directional LSTM model on synthetic data for 
 (a) validation and (b) training sets}
 \label{synthetic_perf_vs_epochs}
 \end{figure}

Figure~\ref{synthetic_perf_vs_epochs} shows the prediction performance of our bi-directional LSTM model in terms of accuracy, precision and recall over the number of epochs. In this model, we have the following settings: embedding size: $3$, LSTM cell: bi-directional, batch-size: $512$, learning rate: $0.02$ and hidden layer size: $6$. Our proposed LSTM model achieves the highest performance after $150$ epochs for both training and validation sets. Table~\ref{table} shows the performance comparison between the proposed LSTM model at epoch $150$, Random Forest algorithm with maximum depth of 4 and 10 trees with bag of words (BoW), and Random Forest algorithm with maximum depth of $4$ and $10$ trees and with no feature engineering.
As can be seen in Table~\ref{table}, our proposed LSTM model outperforms the other two in predicting code failures in our synthetic data. The main reason behind the superior performance of the LSTM model is that it can remember and learn both short-term and long-term characteristics of a sequence, while the other two methods cannot learn the short-term and long-term dependencies. We also tried sequential rule mining method to predict code failures in our synthetic data, but we observed the sequential rule mining fails to fully learn the logic we used to generate the synthetic data and cannot accurately predict the sequences with code failure. This happens because the class of sequential rule mining methods are capable of extracting \textit{positive} rules that result in an outcome (contributors in this context), but they cannot learn and deduct \textit{negative} rules that prevents a specific outcome from happening (blockers in this context). As a result, sequential rule mining methods generally have poor performance in applications that we have a rule which can override and negate the general rule. 

\begin{table*}
\caption {Performance comparison of various code failure prediction algorithms on the synthetic test data}
\label{table}
\begin{center}
\begin{tabular}{|c||c|c|c|} \hline
           & \multicolumn{3} {c|}{\textbf{Metrics}}        \\ \cline{2-4}
\textbf{Prediction Algorithms}  & Precision  & Recall & F1-measure  \\ \hline \hline
RandomForest (Depth=4, Trees =10), No FE          & 0.65  & 0.64  & 0.64    \\ \hline
RandomForest (Depth=4, Trees =10), BoW          & 0.78    & 1.0  & 0.87      \\ \hline
Bi-Directional LSTM (e=3, lstmSize=6)          & 1.0 & 1.0  & 1.0       \\ \hline
\end{tabular}
\end{center}
\end{table*}

In Table~\ref{extraction_sample}, we provide some examples to show how we can find contributors and blockers using algorithm \ref{algo1}. As can be seen in the table, our algorithm is capable of extracting both contributors and extractors correctly for almost all sequences. The only exception is case $7$ where contributors are detected incorrectly.   

\begin{table*}
\caption{Blocker and contributor extraction for a sample of sequences with different length} 
\label{extraction_sample}
\begin{center}
\begin{tabular}{|c|c|c|c|c|c|} \hline
\textbf{id} &\textbf{Sequence} &\textbf{Prediction} &\textbf{True Label} & \textbf{Confidence Score} & \textbf{Comment}\\ \hline
0 & a f b c\colorbox{red}{e}f & 0 & 0 & 1.0 & Correct Extraction\\ \hline
1 & c a f h f c e c k b f a b j e & 0 & 0 & 1.0 & Correct Extraction\\ \hline
2 & a\colorbox{green}{f b c}a  & 1 & 1 & .99 & Correct Extraction\\ \hline
3 & g b g a c a\colorbox{green}{f} b c k\colorbox{green}{b} c f\colorbox{green}{c}  & 1 & 1 & 1.0 & Correct Extraction\\ \hline
4 & g b d f g\colorbox{green}{f} i g \colorbox{green}{b c}   & 1 & 1 & 1.0 & Correct Extraction\\ \hline
5 & \colorbox{green}{f} h a d \colorbox{green}{b} d h f \colorbox{green}{c} g b j d & 1 & 1 & 1.0 & Correct Extraction\\ \hline
6 & k f b c j b h f c \colorbox{green}{f} c \colorbox{green}{b} f \colorbox{green}{c} & 1 & 1 & 1.0 & Correct Extraction\\ \hline
\color{red}{\textbf{7}} & h \colorbox{green}{b j c} a k c d c \colorbox{green}{f b c} i d& 1 & 1 & 0.878 & \color{red}{\textbf{Wrong Extraction}} \\ \hline
8 & \colorbox{green}{f} c d  \colorbox{green}{b} l g l c i \colorbox{green}{c} b f a b& 1 & 1 & 1.0 & Correct Extraction\\ \hline
\end{tabular}
\end{center}
\end{table*}

Sequential rule mining is another set of solutions that can be used for the code failure pattern extraction. We tried a modified version of the algorithm featured in \cite{zaki2001spade} to extract the top rules for our synthetic data. The top rules are listed in Table~\ref{sequential}. Based on the results, this sequential rule mining technique is able to identify the contributors as $(f,b,c)$. However, it fails to detect the blocker event $(e)$. Moreover, this sequential rule mining technique fails to work for our real Excel data where the size of data is much larger and the number of telemetry events is far more than the synthetic data. As another limitation, many recent sequential rule mining algorithms are in fact partially sequential in the sense that in a rule $X \Rightarrow Y$, only $Y$ is required to occur after $X$ and the events in $X$ are unordered \cite{fournier2014fast} \cite{fournier2011rulegrowth}.

\begin{table}
\caption{Top records of sequential rule mining algorithm for the synthetic dataset}
\label{sequential}
\begin{tabular}{|c|c|c|c|} \hline
\textbf{Rules}	& \textbf{Support,\%}	& \textbf{Confidence,\%}	&\textbf{Lift}	\\ \hline						
f,b,c $\Rightarrow$code failure	& 21.55   &45.90	& 2.1304	\\ \hline
b,c$\Rightarrow$code failure	& 21.55 	& 29.176	& 1.3539	\\ \hline
b,f$\Rightarrow$code failure	& 21.55 	& 29.00	& 1.3461	\\ \hline
c,f$\Rightarrow$code failure	& 21.55 	& 28.802	& 1.3365	\\ \hline
b$\Rightarrow$code failure	& 21.55 	& 23.0186	& 1.0681	\\ \hline
c$\Rightarrow$code failure	& 21.55 	& 23.0137	& 1.0679	\\ \hline
\end{tabular}
\end{table}

Table~\ref{table_removing} shows the performance of our proposed algorithm for zero inserting and void inserting approaches.
Based on the results, the void inserting approach outperforms the zero inserting approach. The model is not able to make as accurate predictions using the zero inserting method because there is no sequence in our original dataset that contains $0$ in the middle of the sequence, which means the LSTM network does not have the chance to learn this condition properly during the training process. 

\begin{table*}[t]
\caption {Performance of the proposed LSTM-based method in finding contributors and blockers in synthetic data}
\label{table_removing}
\begin{center}
\begin{tabular}{|c||c|c|c|} \hline
           & \multicolumn{3} {c|}{\textbf{Extraction Accuracy vs. data size}}        \\ \cline{2-4}
\textbf{Algorithms}  		& \textbf{15K}   & \textbf{10K} & \textbf{5K}  \\ \hline \hline
Bi-Dir. LSTM + Zero Inserting         & \begin{tabular}{@{}c@{}} Contributors: Pr=\%78, Rec=\%99 \\ Blockers: Pr=\%80, Rec=\%100\end{tabular} 
									  & \begin{tabular}{@{}c@{}} Contributors: Pr=\%85, Rec=\%83 \\ Blockers: Pr=\%98, Rec=\%99 \end{tabular} 
  									  & \begin{tabular}{@{}c@{}} Contributors: Pr=\%80, Rec= \%80 \\ Blockers: Pr=\%89, Rec=\%92 \end{tabular}    \\ \hline
Bi-Dir. LSTM + Void Inserting         & \begin{tabular}{@{}c@{}} Contributors: Pr=\%99, Rec=\%99 \\ Blockers: Pr=\%99, Rec=\%97 \end{tabular}   
									  & \begin{tabular}{@{}c@{}} Contributors: Pr=\%97, Rec=\%94 \\ Blockers: Pr=\%93, Rec=\%100   \end{tabular} 
									  & \begin{tabular}{@{}c@{}} Contributors: Pr=\%89, Rec=\%89  \\ Blockers: Pr=\%86, Rec=\%100 \end{tabular}      \\ \hline
\end{tabular}
\end{center}
\end{table*}

\begin{figure}[h]
 \centering
 \includegraphics[scale=.55]{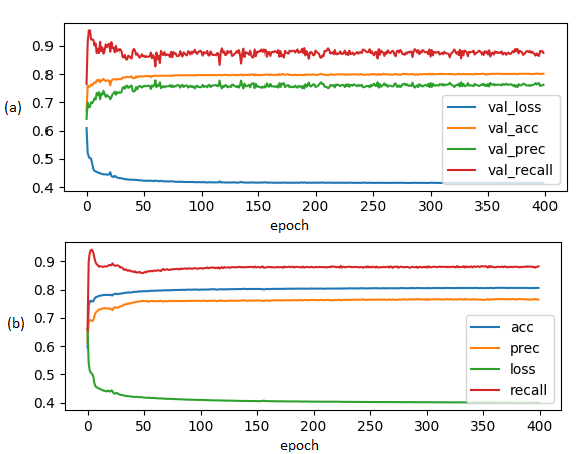}
 \caption{Performance of our bi-directional LSTM model on real data from Excel users for (a) validation and (b) training processes}
 \label{network_prod}
 \end{figure}

\begin{figure}[b]
 \centering
 \includegraphics[scale=.4]{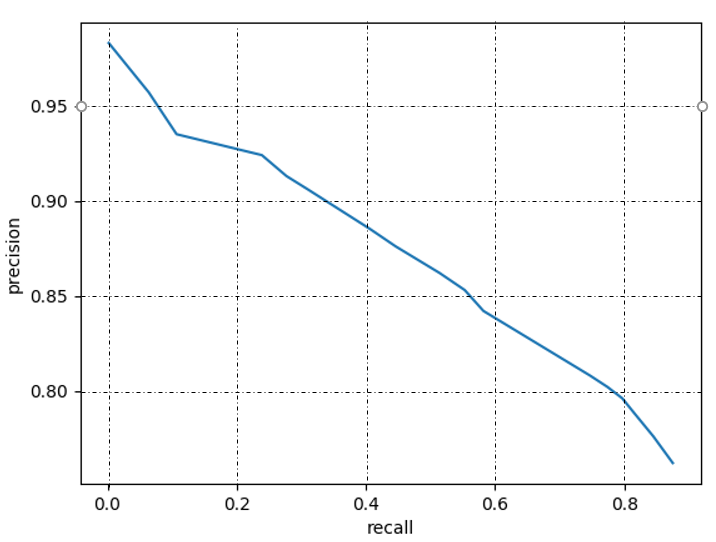}
 \caption{Precision-Recall plot for the proposed bi-directional LSTM network based on real data from Excel users}
 \label{network_prod_pr_rec}
 \end{figure}

Fig.~\ref{network_prod} shows the performance of our LSTM model in terms of accuracy, precision and recall during the training and test process using the real Excel data. As can be inferred from the plot, our trained LSTM model is capable of predicting code failure with over $80$\% accuracy and more than $85$\%
recall. In order to find the root-cause of code failures, we need to focus on  sequences that our LSTM model predicts code failure for them with high confidence. Therefore, we only consider the sequences whose confidence score is above $90$\%. Confidence score is the output of the Dense layer in Fig~\ref{lstm}. As can be seen in Fig.~\ref{network_prod_pr_rec}, our proposed LSTM model has a precision of $90\%$ and above for almost $30$\% of the sequences. 

\section{Practical Remarks: Code Failure Signature Removal}
We need to apply a number of preprocessing steps on the real data before using the proposed LSTM model to predict code failures and extract contributors and blockers. First, we need to remove any event that indicates a signature of the code failure or normal execution of the code. These events are highly correlated with the outcome but do not hold any useful information about the code failure. 
The easiest way to detect such events is to first train our model without any filtering, and extract the contributors and blockers. Then, by showing the results to the subject experts, we can decide whether a specific event is just a signature or it is a real contributor or blocker. For example, in our case we initially identified an event as a blocker which was only called when Excel was properly closed with no code failure. This event was initially picked up by the proposed LSTM model as a blocker, while in reality it was a signature for proper termination of the application.

We also noticed that there are many identical sessions in our real Excel data. In other words, we have many sessions with the same sequence of events. We have two options to address this issue. We can either remove the redundant sessions and train the LSTM model with the reduced dataset, or, keep the original dataset with all redundant data. It is clear that models developed from these two approaches would be quite different. To solve this delima, it is recommended to make sure that the training data is a true representation of the real data and mimic the real-world conditions as closely as possible. Therefore, removing the redundant data points is not a good option in our code failure prediction example.

\section{Conclusion}
In this paper, we applied LSTM recurrent neural networks to find sessions that are prone to code failure and to extract telemetry patterns that lead to a specific code failure. Our method is designed to process a large set of data and automatically handle edge cases in code failure prediction. We took advantage of Bayesian optimization technique to find the optimal hyper parameters. To extract the failue code patterns, we first introduced the Contributors and Blockers concepts and we used a greedy approach to find them. 
We used both synthetic and real data to develop and test our proposed LSTM model. Our trained LSTM model demonstrated
over $99\%$ accuracy for detecting code failures in the synthetic data. Using the proposed greedy method, we detected the contributors and blockers in the synthetic data in more than $90\%$ of the cases, with a performance better than sequential rule and pattern mining algorithms. 
\section{Acknowledgment}
The author would like to thank Dr. Sandi Ganguli and Wayne Roseberry for constructive discussion and providing experimental data.
\bibliographystyle{IEEEtran} 
\bibliography{IEEEabrv,seqdb_lstm}
\end{document}